\documentclass[letterpaper, 10pt, conference]{ieeeconf}      

\IEEEoverridecommandlockouts                              

\usepackage{graphicx} 

\usepackage{epsfig} 
\usepackage{epstopdf}
\usepackage{tikz}
\usepackage{mathptmx} 
\usepackage{times} 
\usepackage{amsmath} 
\usepackage{amssymb}  

\usepackage{algorithm}

\usepackage{tablefootnote}
\usepackage{algpseudocode}
\usepackage{color}
\usepackage{amsfonts}
\usepackage[noadjust]{cite}
\usepackage{subfigure}
\usepackage{threeparttable}
\usepackage[framed,numbered,autolinebreaks,useliterate]{mcode}
\usepackage{soul}
\usepackage[normalem]{ulem}
\usepackage{pgf}
\usepackage{tikz}
\usetikzlibrary{arrows,automata}
\usepackage{ifthen}
\usepackage{booktabs}

\usepackage[a-1b]{pdfx}

\usepackage{hyperref} 

\usepackage{todonotes}

\makeatletter
\newcommand\fs@spaceruled{\def\@fs@cfont{\bfseries}\let\@fs@capt\floatc@ruled
  \def\@fs@pre{\vspace{0.5\baselineskip}\hrule height.8pt depth0pt \kern2pt}%
  \def\@fs@post{\kern2pt\hrule\vspace{-0.95\baselineskip}}
  \def\@fs@mid{\kern2pt\hrule\kern2pt}%
  \let\@fs@iftopcapt\iftrue}
\makeatother

\title{\LARGE \bf Augmented-Reality Enabled Crop Monitoring with Robot Assistance} 

\author{Caio Mucchiani, Dimitrios Chatziparaschis, Konstantinos Karydis
\thanks{$^{1}$ Dept. of Electrical and Computer Eng., Univ. of California, Riverside, 900 University Avenue, Riverside, CA 92521, USA. Email: {\tt\footnotesize\{caiocesr, dchat013, karydis\}@ucr.edu}. 
We gratefully acknowledge the support of NSF \#CMMI-2046270 and \#CMMI-2326309, USDA-NIFA \#2021-67022-33453, ONR \#W911NF-22-1-0156, and The University of California under grant UC-MRPI M21PR3417. 
Any opinions, findings, and conclusions or recommendations expressed in this material are those of the authors and do not necessarily reflect the views of the funding agencies.
}}

\begin{document}
\maketitle

\begin{abstract}
The integration of augmented reality (AR), extended reality (XR), and virtual reality (VR) technologies in agriculture has shown significant promise in enhancing various agricultural practices. Mobile robots have also been adopted as assessment tools in precision agriculture, improving economic efficiency and productivity, and minimizing undesired effects such as weeds and pests. Despite considerable work on both fronts, the combination of a versatile User Interface (UI) provided by an AR headset with the integration and direct interaction and control of a mobile field robot has not yet been fully explored or standardized. This work aims to address this gap by providing real-time data input and control output of a mobile robot for precision agriculture through a virtual environment enabled by an AR headset interface. The system leverages open-source computational tools and off-the-shelf hardware for effective integration. Distinctive case studies are presented where growers or technicians can interact with a legged robot via an AR headset and a UI. Users can teleoperate the robot to gather information in an area of interest, request real-time graphed status of an area, or have the robot autonomously navigate to selected areas for measurement updates. The proposed system utilizes a custom local navigation method with a fixed holographic coordinate system in combination with QR codes. This step toward fusing AR and robotics in agriculture aims to provide practical solutions for real-time data management and control enabled by human-robot interaction. The implementation can be extended to various robot applications in agriculture and beyond, promoting a unified framework for on-demand and autonomous robot operation in the field.
\end{abstract}


\section{Introduction}

The integration of augmented reality (AR), extended reality (XR), and virtual reality (VR) technologies in agriculture has shown significant promise in enhancing various agricultural practices. 
These technologies offer innovative solutions for improving efficiency, access to information, and decision-making processes in the field. 
Previous work considered applications such as supervision of multiple agricultural machines for improved situational awareness and machine management~\cite{huuskonen2019augmented},  real-time insect identification and pest management~\cite{nigam2011augmented}, crop information and productivity~\cite{katsaros2017farmar}, plant disease detection~\cite{ponnusamy2021iot}, and yield and grade prediction~\cite{bv2024aria}.
Additional examples include support in farm management activities~\cite{xi2018future}, training in hydroponic agriculture~\cite{covarrubias2024empowering}, and education in the agronomy domain~\cite{greig2024enhancing}, among others~\cite{anastasiou2023applications}. 

Smartphones and tablets are often adopted as user interfaces for these applications. However, the use of mixed reality glasses and AR headsets can improve user experience while enabling additional layers of interaction. 
These range from visualization, remote assistance, and hands-free operations in livestock farming~\cite{caria2019exploring}, field navigation for soil condition and analysis~\cite{zheng2019location}, measurement of understorey vegetation~\cite{gorczynski2022measuring}, and determining fruit weight~\cite{van2022determining}. 
Similarly, the adoption of mobile robots as assessment tools in precision agriculture has already established itself as an important factor in improving the practice economically~\cite{biswas2022improving}, improving productivity~\cite{biswas2022improving}, and minimizing undesired effects such as weeds~\cite{visentin2023mixed} or pests~\cite{gonzalez2017fleets}.  
A summary of AR-enabled applications in agriculture is presented in Table~\ref{tab:summary}, and a proposed taxonomy can be found in~\cite{suzuki2022augmented}.

\begin{figure}[!t]
    \vspace{6pt}
    \centering
    \includegraphics[width=0.99\columnwidth]{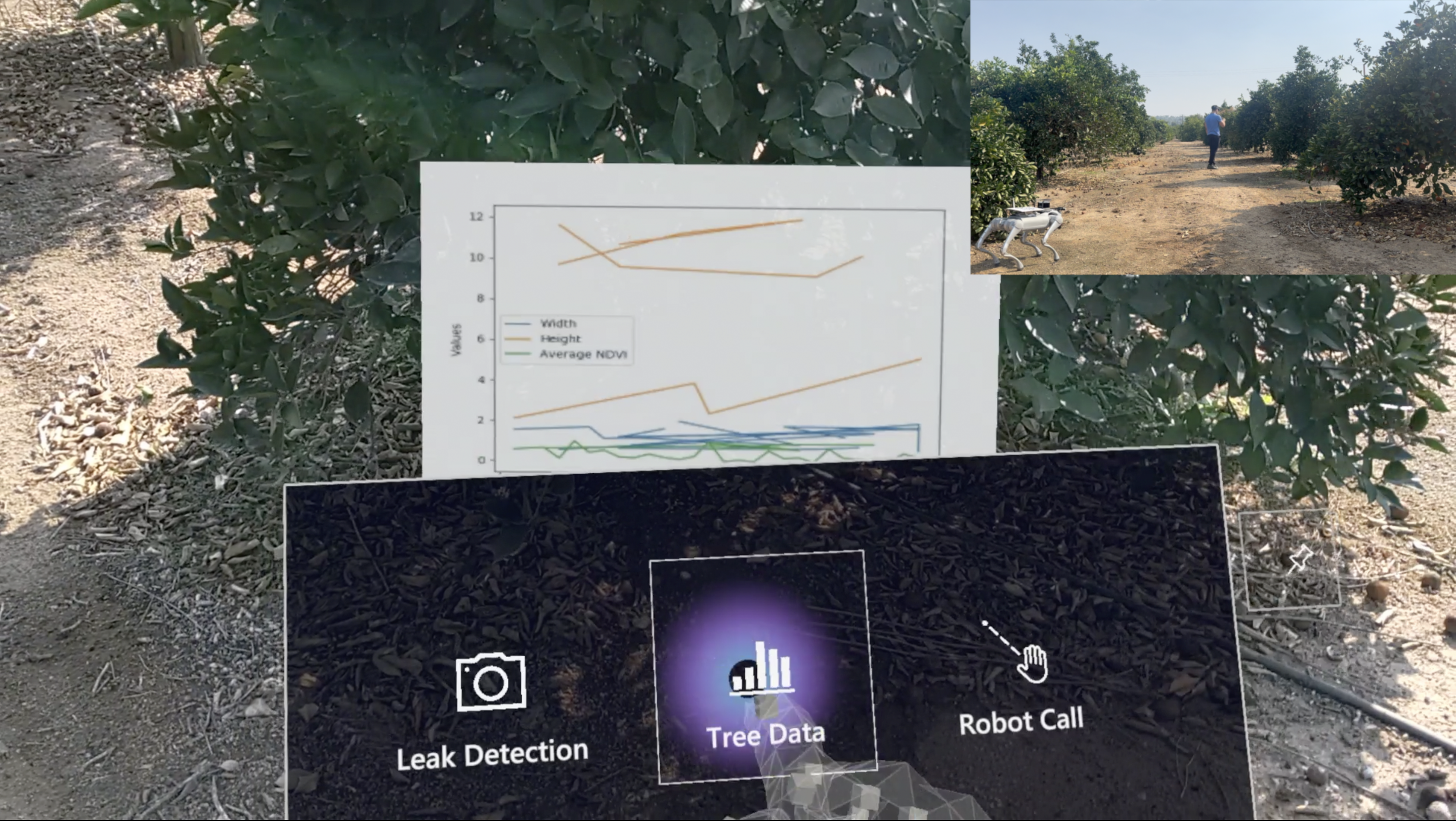}
    \vspace{-12pt}
    \caption{First-person view of the developed AR interface providing real-time crop status and updates with the assistance of a field robot.}
    \label{fig:intro}
    \vspace{-12pt}
\end{figure}

Despite considerable work on both fronts, combining the advantages of a versatile UI such as the one provided by an AR headset, with integration and direct interaction and control of a mobile robot has not yet been explored or standardized, with most work to date providing users with information input only access to the robot's location~\cite{huuskonen2019augmented} or assisting with the user's navigation~\cite{huuskonen2018soil}. 
Previous demonstration proposed the integration of these various components from an AR interface to robot control~\cite{zhang2023accessible} but was not directly applied to a specific domain such as the one proposed herein. 
Further, the different layers of system integration required to create an AR environment (generally designed in Unity) and interface it with the Robot Operating System (ROS) can be a quite complicated process, regardless of the tools already available for such integration~\cite{allspaw2023comparing,zea2021iviz}.

\begin{table*}[!t]
    \vspace{6pt}
    \centering
    \caption{References on Augmented Reality in Agriculture.}
    \label{tab:summary}
    \vspace{-6pt}
\scalebox{0.75}{
\begin{tabular}{l l l l l}
Title & Topic & Type of Device Used & Type of Robot Used & Citation \\ \toprule
Supporting table grape berry thinning with deep neural network and augmented reality technologies & Fruit cultivation & AR Glasses & None & \cite{buayai2023supporting} \\
Augmented reality for supervising multi-robot system in agricultural field operation & Field operation & AR Glasses & Ground Robot & \cite{huuskonen2019augmented} \\
Digital Twins with Application of AR and VR in Livestock Instructions & Livestock education & AR Glasses, Smartphone, Tablet & Haptic Device &  \cite{petrov2021digital} \\
Soil sampling with drones and augmented reality in precision agriculture & Soil sampling & AR Glasses & Aerial Robot & \cite{huuskonen2018soil}\\
Smart pest management: an augmented reality-based approach for an organic cultivation & Pest control & Smartphone, Tablet & None & \cite{mahenthiran2021smart} \\
Research on Training Pilots of Agriculture and Forestry Protection Drone by MR Technology & Forestry protection & MR Glasses & Aerial Robot & \cite{xu2021research} \\
Augmented reality greenhouse & Greenhouse management & Smartphone, Tablet & None & \cite{de2013augmented} \\ \bottomrule
\end{tabular}}
\vspace{-12pt}
\end{table*}

This work aims to address this gap by providing real-time data input and control output of a mobile robot for precision agriculture through a virtual environment enabled by an AR headset interface. 
Our system leverages open-source computational tools and off-the-shelf hardware to promote direct and effective system integration. 
Specifically, we present a case study considering an application for growers or technicians walking along a field (Fig.~\ref{fig:intro}), in which the user can interact with the (legged) robot via an AR headset and a user interface. 
In this application, the user can (1) access the robot's camera and teleoperate it to check for leaks in irrigation lines, (2) request a real-time graphed status of a specific tree (such as dimensions or vegetation indices), and (3) have the robot update measurements, also in real-time, by selecting a desired tree or area and having it autonomously navigate to it. To realize this, we also propose a local navigation method utilizing a fixed holographic coordinate system with moving frames attached to QR codes. Our contributions include developing a real-time AR interface for precision agriculture, implementing AR-based autonomous navigation, and creating the ``Holoagro'' application for farmers to interact with and teleoperate robots for various agricultural tasks. These contributions can advance the integration of AR and robotics in agriculture, providing a practical solution for real-time data management and control enabled by human-robot interaction.
%

\section{System Description}
\label{sec:problem}

Tree dimensions such as height and crown width can indicate tree growth, shading, carbon sequestration, and drought stress, among others~\cite{pretzsch2015crown,grote2016importance}. 
In addition, the importance of vegetation indices such as the Normalized Difference Vegetation Index (NDVI) for agriculture has been increasingly associated with drought and yield variability predictions in certain crops~\cite{jena2019normalized,liakos2015use,meneses2011ndvi}. 
Although each of these indicators alone is not sufficient to precisely determine 
these metrics, a combination of these measures may offer more effective information to growers and field technicians. 
In this sense, and aware of the development of numerous works employing machine learning and robotics in agriculture~\cite{wakchaure2023application}, concurrently with AR (Table~\ref{tab:summary}), we propose a combination of the two as an initial exploration toward effective human-robot interaction (HRI) systems for precision agriculture. 
By leveraging the multimodal sensing ability of mobile robots and the versatile user interaction experience provided by AR, our system enables the user to receive real-time data as collected by the robotic platform in the field. 

\begin{figure}[!t]
   \vspace{6pt}
   \centering
   \includegraphics[width=0.99\columnwidth]{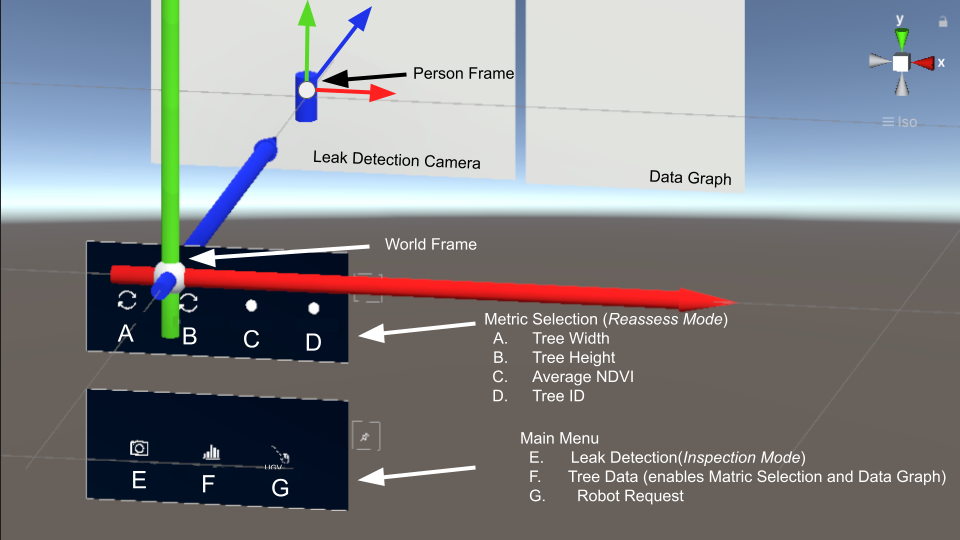}
   \vspace{-12pt}
   \caption{Snapshot from the UI implemented in Unity. A fixed reference frame (World frame) is initiated upon starting the application, while a moving frame at an offset from the user is defined. Two screens are utilized for streaming the robot's front-facing camera and plotting a data graph. The menu allows the user to enable or disable both camera and graph views, while also calling the robot. In the latter case, a secondary menu also pops up for the selection of the desired data to be reassessed.}
   \label{fig:interface}
   \vspace{-12pt}
\end{figure}

Our approach also enables the user to request a robot to resample a specific tree (or area more broadly) when needed, based on their own experience from analyzing the provided data. 
The robot will autonomously navigate toward the selected area and reassess the requested metrics. 
Further, the user can take over control of the robotic platform and, for instance, navigate it around the field when checking for water leakage in irrigation lines, while receiving real-time video feed on the AR interface. 
Our system UI is depicted in Fig.~\ref{fig:interface}; a detailed description of its implementation follows.

\subsection{Hardware Implementation}
\label{sec:hardware}

We used Microsoft Hololens 2 as the AR interface. 
The Hololens has a see-through display with a holographic density of 2.5K radiants (light points per radian), uses a Qualcomm Snapdragon 850 processor with built-in spatial sound and m5-channel microphone array. 
It includes four visible light cameras for head tracking, two infrared cameras for eye tracking, an 8-megapixel (MP) Camera, a 1-MP Time-of-Flight (TOF) depth sensor, and an IMU. 
The field of view (approximately 52 degrees), although limited for immersive sensation, provides sufficient screen space for displaying various UIs perceived as hologram elements. 
Network connectivity with the device can be enabled either through Wi-Fi or USB-C connections.

Regarding accuracy and repeatability, previous work demonstrated the Hololens 2 had recordings of about $25\;mm$ for accuracy and between $5.9$ and $7.2\;mm$ for repeatability when the tracked handheld object was held in the center of the field of vision~\cite{soares2021accuracy}. 
It is concluded these values can be influenced by the tracked object's position in the projected vision field.
These numbers, albeit somewhat large for tasks such as fine manipulation, are within the desired operational envelope of this work.  In addition, the access to various software tools (described next) was a motivating factor for this choice of interface.

Our study also considered a physical legged robot, due to its robustness in traversing unstructured environments such as agricultural fields. 
We have selected the Unitree Go2 for demonstration of our system application. 
The robot has a maximum running speed of 5$m/s$, up to 12$kg$ payload capacity, a 4D Lidar (able to measure speed and detect depth) for autonomous navigation, and a maximum climbing angle of 40 degrees. 
The Go2 utilizes an 8-core CPU and allows connectivity via Wi-Fi, Bluetooth, and 4G. 
Our selection criteria also considered the availability of computational tools enabling integration with robot control via the AR interface. 
A wireless joystick was used for teleoperation by the user when the robot was set to perform inspection tasks. 
A Netgear Nighthawk M6 router enabled a wireless connection between all aforementioned devices. 

\begin{figure*}[!t]
   \vspace{6pt}
   \centering
     \includegraphics[width=0.90\textwidth]{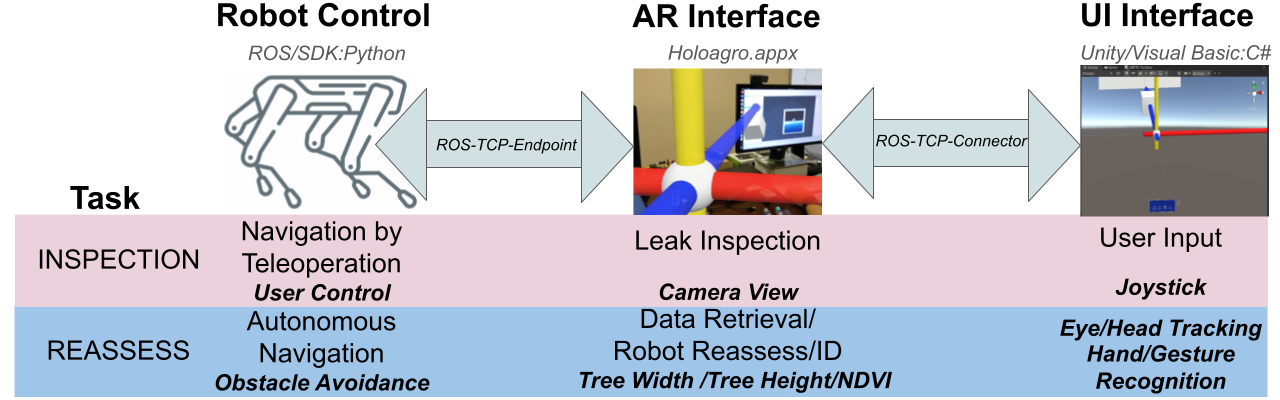}
     \vspace{-3pt}
     \caption{Overview of the implementation fusing the robot, AR, and UI interfaces. The communication between ROS and Unity is facilitated by the ROS-TCP-Connector and ROS-TCP-Endpoint modules. Different modalities for the system implementation are input by joystick, eye and head tracking, and hand or gesture recognition. The tasks considering these inputs are referred to as inspection and reassess tasks.}
     \label{fig:pipeline}
     \vspace{-6pt}
\end{figure*}

\subsection{Software Implementation}
\label{sec:software}

The UI on the Hololens side was implemented in Unity, a game engine for cross-platform development widely used for VR and AR environment creation. 
Software development utilized the Mixed Reality Toolkit (MRTK)~\cite{mrtk}, which is also cross-platform, 
and the OpenXR API to enable interaction profiles (i.e. a set of tools) such as hand and eye tracking. 

For software development on the legged robot, we used the official Unitree Python SDK 2~\cite{sdk} along with ROS. 
To permit communication between the devices, two Unity packages, ROS-TCP-Connector and ROS-TCP-Endpoint were utilized. 
By connecting all devices to the same network, these packages can transmit and receive ROS topics, messages, and services from and to each other (Fig.~\ref{fig:pipeline}). 
This approach has been found to work well in related past works on Unity-based systems~\cite{zhang2023accessible,sheng2023design}. 
Other open-source implementations were also considered to generate poses for QR code detection~\cite{qrcodedetector} and camera feedback~\cite{zhang2023accessible}.

We break down our implementation into modalities for the user interaction, AR Interface, and the legged robot motion (Fig.~\ref{fig:pipeline}). 
We considered two levels of interaction with the robot: teleoperation which relates to inspection tasks and autonomous navigation which applies to ``reassess" tasks. 
The former requires the user to teleoperate the robot using the AR interface along with a joystick to achieve a task of interest.\;\footnote{~The task considered herein was leak detection in an irrigation line, but other similar tasks are essentially possible as well.} 
The latter transmits a request to the robot to reach a pose specified by the user via the AR UI; once the robot arrives, it reassesses any parameters of interest (as selected by the user) there (for instance tree height).\;\footnote{~The choice to provide information as requested by the user was made so that only relevant information at a given time is provided by the user, and it is not a limiting factor. 
In fact, the robot can still collect additional information based on the onboard sensors (e.g., a multi-modal scene understanding setup~\cite{teng2023multimodal}) and fuse multi-modal information as it navigates~\cite{chatziparaschis2024go} to update any crop database information.} 

\section{Local Navigation}
\label{sec:localnav}

\begin{figure}[!t]
   \vspace{6pt}
   \centering
   \includegraphics[width=1\columnwidth]{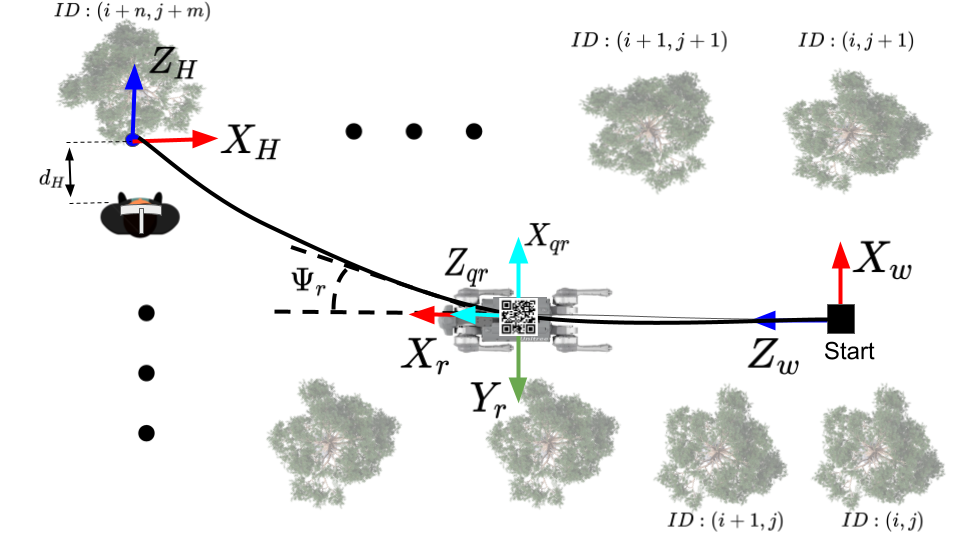}
     \vspace{-18pt}
      \caption{Coordinate systems defined for the local navigation problem. The virtual environment frames, which include the origin, QR code, and goal poses, follow a left-hand side convention while the robot's local reference uses the right-hand convention.}
      \label{fig:frames}
      \vspace{-12pt}
\end{figure}

To determine a goal pose for a robot using AR in a holographic coordinate system, we need to consider three different reference frames (Fig.~\ref{fig:frames}) under the same global coordinate system. 
By defining the (holographic) world frame $W$, we have the AR interface frame $H$ as a moving frame, under which we create the goal attached at an offset $d_H$ from the $H$ origin (and therefore the user/operator that uses it). 
To obtain the operator's global position in $W$, we use the transformation matrix $_{W}^{H}T$.

On the robot side, we define as $r$ the robot's moving frame, provided via odometry described by its started origin, and we place a fixed QR code on its body frame at a known position, described by a static transformation $_{r}^{qr}T$. 
At the start of the experiment and (if desired) in the latter moments during the survey, we use the Hololens headset to detect and localize the QR code in the holographic frame $W$, thus instantly obtaining the robot's global pose via the transformation $_{W}^{r}T~=~_{W}^{qr}T \cdot ~_{r}^{qr}T^{-1}$. 
Creating a frame attached to the robot in this manner facilitates measurements in the holographic world. Despite the addition of one more step, i.e. to scan the robot's QR code during system initialization, it enables a local navigation method and minimizes the need to integrate additional costly hardware for positioning (e.g., RTK-GPS) while taking advantage of the AR headset capability to generate coordinate systems in the real world through holograms. 
As long as the robot is operating in the field, odometry pose changes are added to the captured transformation $_{W}^{r}T$ to keep track of the robot's global position in the holographic frame $W$, even if it is not under the human operator field of view. 

Our local navigation method can be embedded into a 2D setting using waypoints. 
Waypoint creation can be expressed in terms of a plane parallel to the physical ground and a rotation about an axis perpendicular to that plane to enable the robot to orient toward the goal.  
However, the coordinate system conventions between Unity and ROS are distinct, hence a static transformation must be integrated as well. 
Specifically, the former uses a left-handed rule and clockwise positive rotation with the $Y$ axis up, while ROS adopts the right-hand rule with counter-clockwise rotation and the $Z$ axis up. 
The relation between the two is given by 
\begin{equation}
[X,Y,Z,q_x,q_y,q_z,q_w]_{Unity} = [Z,-X,Y,-q_z,q_x,q_y,q_w]_{ROS}\;.
\end{equation}
where $[X,Y,Z]$ denotes position and $[q_x,q_y,q_z,q_w]$ orientation in quaternions representation. 

The robot's own frame also considers the right-hand convention, with the $X$ axis pointing forward and counter-clockwise positive rotation. 
We define the heading angle $\Psi_H$ between the robot (in holographic coordinates) and the goal, and a distance $d_h=0.5\;m$ as to offset the goal from the user's position. 
To correlate tree positions with a desired goal in the environment, we assumed a known prior map identifying and labeling trees as $ID:(i,j)$ from the southeasternmost one to $ID:(i+m,j+n)$ as the northwestmost tree in the field. 
Therefore, the desired pose of the robot at tree $ID:(i,j)$, in the holographic world coordinate system is given by
\begin{equation}
    P_d(i,j) = [x_H,z_H,\psi_H]_W^r\;.
\end{equation}

These waypoints are set by the user via the AR UI and can be placed at any time, with the robot dynamically adapting its trajectory to the destination using its built-in obstacle avoidance capabilities.\;\footnote{~Our framework can apply directly to any robot that has built-in obstacle avoidance capabilities. Otherwise, autonomous navigation methods involving environment perception and planning in agricultural settings (e.g.,~\cite{zhang2020high,dechemi2023robotic,lee2024towards}) need to be integrated into our developed framework as well.}

\section{Field Deployment and Testing}
\label{sec:deployment}


We consider two example tasks to illustrate the developed ``Holoagro'' application for the AR-robot integrated system. 
The first example (an inspection task) enables the user to teleoperate robot and access its onboard camera feed through the dedicated UI, in real-time. 
The second example (a reassess task) provides the user with the ability to visualize real-time field data, specific to a tree, and request robot reassessment. These tasks intend to demonstrate the system's integration between ROS and the AR application, rendering a bi-directional communication means between a virtually created environment and the physical one, via a fusion of visual and physical feedback.

\subsection{Modes of Interaction}
Each suggested mode requires user input and interaction with the robot and AR interface at different levels. 

\begin{itemize}
    \item AR Interface: The interface consists of camera feedback (e.g., to enable a user to assess the presence of leaks in irrigation lines during an inspection task), graphed data based on corresponding user location, and a robot callback utility to reassess data for the Reassess task.
    \item User: Input is collected from either eye or head tracking, as well as hand or gesture recognition. The AR interface then interprets these inputs and matches the corresponding action with the designed hologram. Direct input to the robot via the joystick is enabled as well.
    \item Legged Robot: Teleoperation for inspection tasks is enabled via the joystick, while autonomous motion with obstacle avoidance is enabled during the reassess tasks.
\end{itemize}

\begin{figure}[!t]
   \vspace{6pt}
   \centering
   \includegraphics[trim={0cm, 0cm, 0cm, 0cm},clip,width=1\columnwidth]{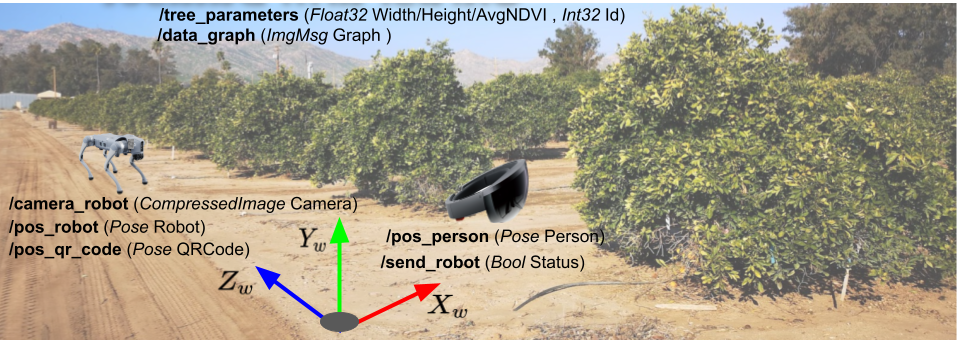}
     \vspace{-15pt}
      \caption{Summary of ROS topics published by the AR application. The world frame is depicted and created when initializing the application.}
      \label{fig:topic}
      \vspace{-15pt}
\end{figure}

To collect the desired poses, determine user requests, and broadcast appropriate camera views, various ROS topics are published and shared between the robot and the AR interface via the ROS-TCP package. 
These are visualized in Fig.~\ref{fig:topic} and include the following.

\begin{itemize}
    \item \texttt{camera\_robot}: \textit{sensor\_msgs/CompressedImage} message type topic publishing a live stream of the robot's front-facing camera during inspection tasks.
    \item \texttt{pos\_robot}: \textit{geometry\_msgs/Pose} message type topic publishing the pose of the robot with respect to the world frame. 
    \item \texttt{pos\_qr\_code}: \textit{geometry\_msgs/Pose} message type topic also publishing the pose of the robot via the QR code on it with respect to the world frame.
    \item \texttt{pos\_person}: \textit{geometry\_msgs/Pose} message type topic publishing the pose of the user via the AR headset with respect to the world frame.
    \item \texttt{send\_robot}: \textit{std\_msgs/Bool} message type topic publishing the request to enter in reassess task mode.
    \item \texttt{tree\_parameters}: \textit{std\_msgs/Float32,std\_msgs/Int32} messages type topic publishing tree parameters such as width, height, and average NDVI per tree ID.
    \item \texttt{data\_graph}: \textit{sensor\_msgs/Image} message type topic publishing a real-time updated graph of the tree parameters for all IDs registered in the given map.
\end{itemize}

We demonstrate our system at a citrus tree field located at the Agricultural Experimental Station at the University of California, Riverside (AES; $33^\circ{}~58'~3.2592''~N,~117^\circ{} 20'~7.0296'' W$). 

\subsection{Performing an Inspection Task}
\begin{figure}[!t]
   \vspace{6pt}
   \centering     \includegraphics[width=1\columnwidth]{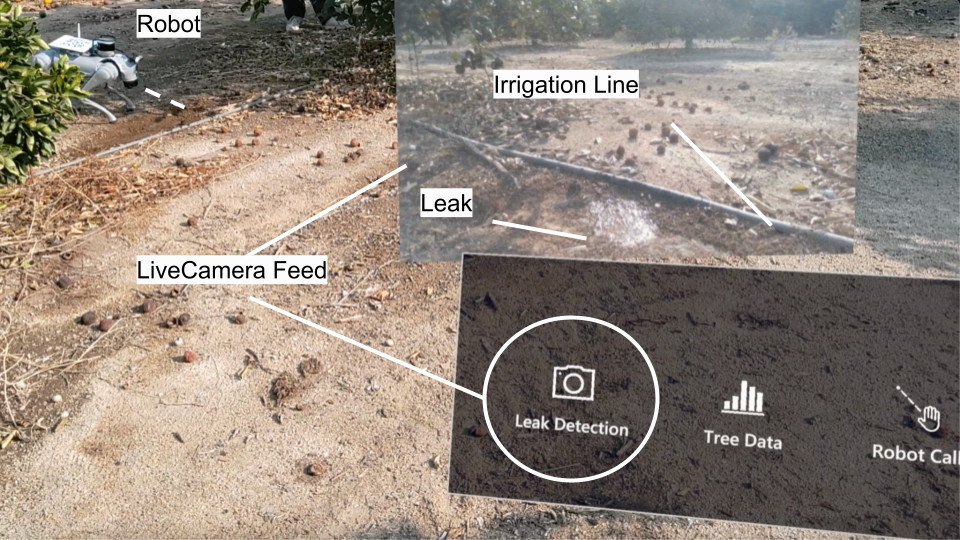}
     \vspace{-12pt}
      \caption{First-person view of the AR UI during an inspection task (here detection of an irrigation pipe leak). The same UI can be used when in other tasks since camera feedback and teleoperation can be activated on demand.}
      \label{fig:fov}
      \vspace{-12pt}
\end{figure}

During inspection task operation mode a window appears in the upper right corner of the UI (Fig.~\ref{fig:fov}) and live streams the feed from the robot’s front-facing camera. 
The user can teleoperate the robot based on the camera feed to check the surroundings.
In the specific sample case study considered herein, the user looks for any leaks in irrigation lines. 
As a result, our framework enables the user to clearly spot a leak by changing the posture of the robot. 
Although, at the current stage of development, the robot's motion is directly controlled by the user via a joystick, our framework affords integration into the UI directly through the SDK, so that the user can control all motion within the virtual interface. 



\subsection{Performing a Reassess Task}
\begin{figure*}[!t]
   \vspace{6pt}
   \centering
     \includegraphics[trim={0cm, 0cm, 0.10cm, 0cm},clip,width=\textwidth]{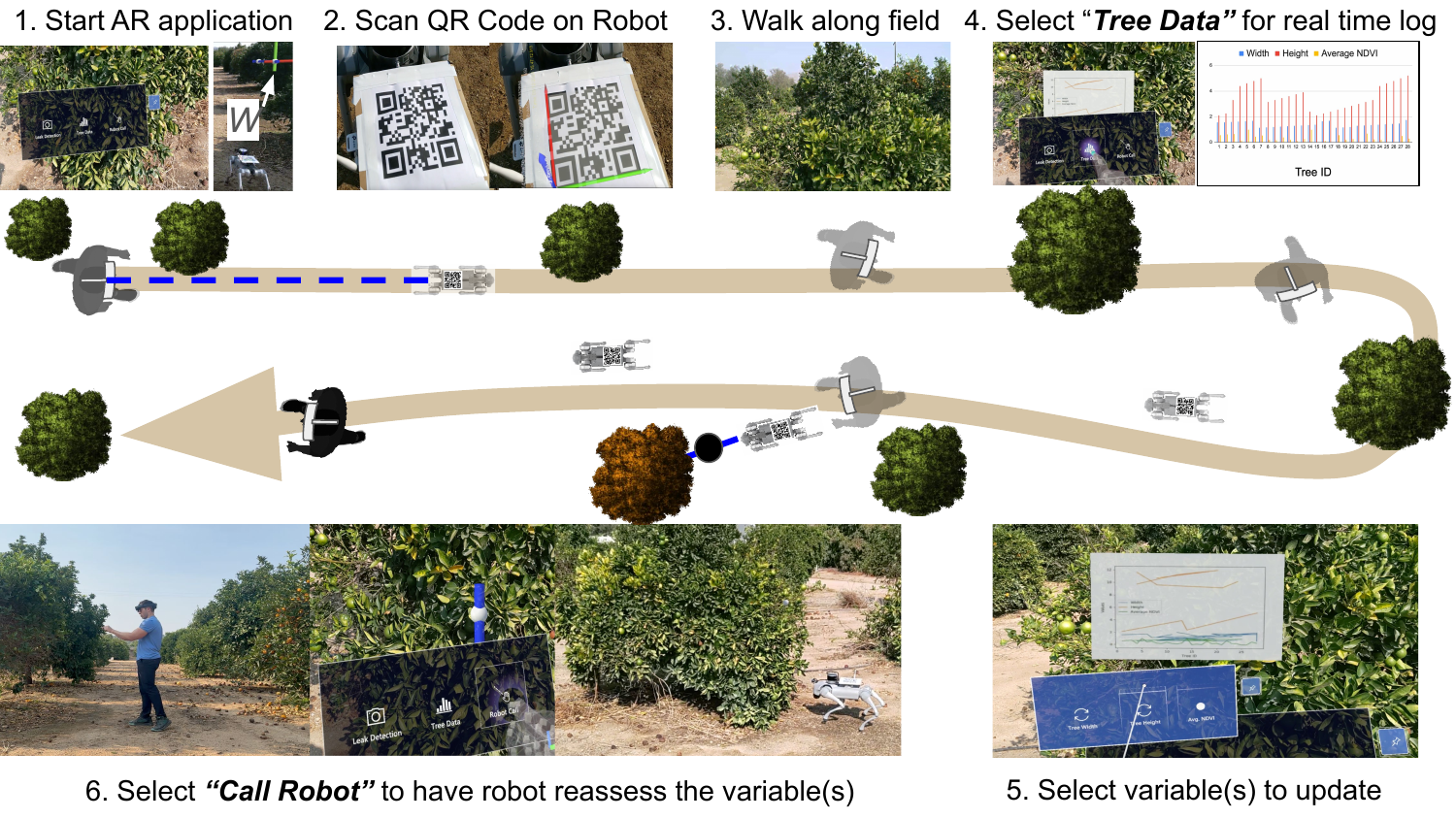}
     \vspace{-18pt}
      \caption{Overview of the system pipeline during reassess task operation mode. (Best viewed in color.) }
      \label{fig:overview_reassess}
      \vspace{-15pt}
\end{figure*}

A series of steps occurs while in reassess task operation mode. 
Figure~\ref{fig:overview_reassess} visualizes the general process. 
Upon starting the application, a world frame $W$ is created and placed at the AR headset current pose. 
The user then proceeds toward the robot, scanning its QR code, such that the application also creates a reference frame at that pose for local navigation.
The user then walks along the field checking tree conditions when deciding which variable(s) to update. 
Once selected, the user can call the robot to do a reassessment of any or all the chosen data.

\floatstyle{spaceruled}
\restylefloat{algorithm}
\begin{algorithm}[!t]
\caption{Robot Navigation using AR and QR Code}\label{alg:robot_navigation}
\begin{algorithmic}[1]
    \State \textbf{Subscribe to ROS Topics from AR headset:}
    \State Subscribe to \texttt{pos\_person} and \texttt{qr\_code\_pos}.
    \State Publish to \texttt{tree\_parameters} and \texttt{data\_graph}.
    \State \textbf{Retrieve Coordinates:}
    \State \( x_h, z_h,\psi_h \leftarrow \text{AR glasses} \)
    \State \( x_{qr}, z_{qr}, \psi_{qr} \leftarrow \text{QR code} \)
    
    \State \textbf{Save Robot Local Pose as Offset:}
    \State \( x_r, y_r, \psi_r \leftarrow \text{robot initial local pose} \)

   \Function{local\_to\_global}{$x_r, y_r, \psi_r, x_{qr}, z_{qr}, \psi_{qr}$}
        \State Scan QR code on the robot.
        \State Apply 2D static transformation.
        \State \Return \( x_{r\_new}, z_{r\_new}, \psi_{r\_new} \)
    \EndFunction
    \While{True}
    \State \textbf{If New Goal Position is Received:}
    \State \( \psi_{\text{goal}} \leftarrow \text{relative\_yaw}(x_{r\_new}, z_{r\_new}, \psi_{r\_new}, x_h, z_h) \)
    \State \( d_{\text{goal}} \leftarrow \text{relative\_distance}(x_{r\_new}, z_{r\_new}, x_h, z_h) \)
    
    \While{$|\psi_{\text{goal}}| \geq 2^\circ$}
        \State Rotate \( \omega=0.4 \; sign(\psi_{goal}) \; rad/s\)
    \EndWhile
    
    \While{$|d_{\text{goal}}| \geq 0.1$ m}
        \State Move Forward $v = $\( 0.6 \) $m/s$
    \EndWhile
    
    \State \textbf{Publish} \texttt{tree\_parameters} \textbf{for} $Tree(i,j)$:
        \State $\texttt{tree\_parameters.W} \leftarrow Tree(i,j).W$
        \State $\texttt{tree\_parameters.H} \leftarrow Tree(i,j).H$
        \State $\texttt{tree\_parameters.NDVI}  \leftarrow Tree(i,j).NDVI$
    \State \textbf{Publish} \texttt{data\_graph} \textbf{for} $Tree(i,j)$:
        \State $\texttt{data\_graph} \leftarrow  plot(Tree(i,j))$
    \EndWhile
\end{algorithmic}
\end{algorithm}

The software implementation of the developed robot callback function is detailed in Alg.~\ref{alg:robot_navigation}. 
We obtain the goal position through the goal (\texttt{pos\_person}) and the QR code pose (\texttt{qr\_code\_pos})  as ROS topics, while publishing the \texttt{tree\_parameters} and \texttt{data\_graph} for the chosen tree (lines 1 to 3). 
By acquiring the AR headset and QR code poses in $W$ coordinates, we first save the robot's local pose (line 7), convert it to a global ($W$) pose, and calculate its relative angle and distance to the goal. 
The robot autonomously moves toward the goal with constant angular and linear velocities (determined by the user; herein set at $0.4\;rad/s$ and $0.6\;m/s$ respectively) while avoiding obstacles up until within a user-defined offset from the goal (herein set at $0.035\;rad$ in orientation and $0.1\;m$ in translation). 
At the goal pose, the robot collects information from its sensors and updates the selected parameters of the tree, which are here its \textit{width, height}, and average canopy \textit{NDVI}, based on earlier work~\cite{chatziparaschis2024go}. 
These parameters, along with the updated graph, are published in ROS while the robot awaits a new goal pose from the user. 
Our framework can handle successive goal poses, allowing the user to specify multiple targets while moving through the field. 

\begin{figure}
    \centering
    \includegraphics[trim={1.70cm, 0cm, 2.35cm, 1cm},clip,width=\columnwidth]{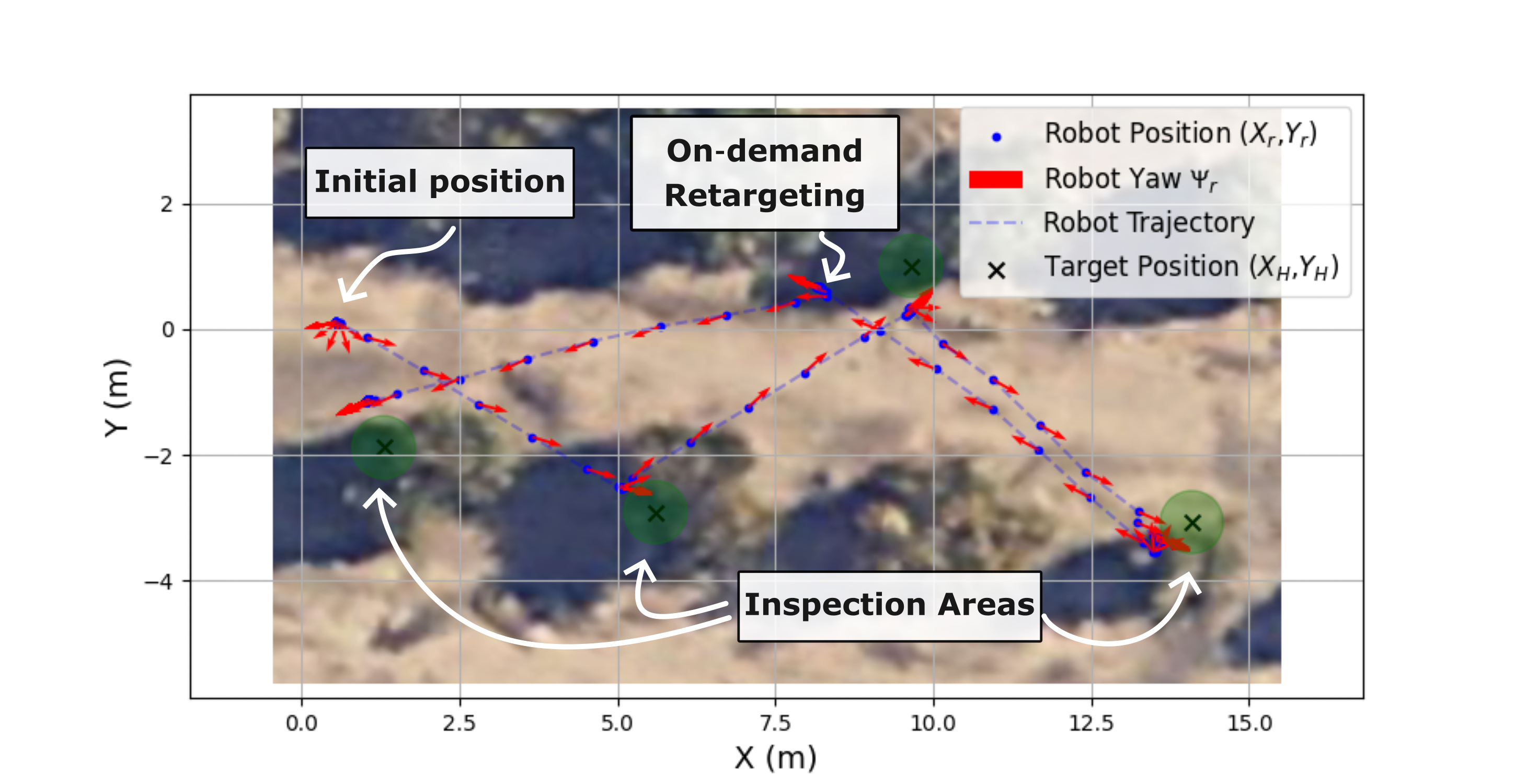}
    \caption{Robot behavior during the inspection of a citrus orchard. The operator provides areas of interest for the robot to check during the inspection, depicted as green areas $(X_H, Y_H)$ with $d_H$, and the robot   $(X_r, Y_r, \Psi_r)$ can autonomously navigate toward them. At all times the robot keeps a safe distance from surrounding objects, especially when reaching the desired tree, and is able to readjust to a newly obtained target. (Best viewed in color.)}
    \label{fig:fourtreeexperiment}
    \vspace{-12pt}
\end{figure}

Figure~\ref{fig:fourtreeexperiment} demonstrates a scenario using the robot in a reassess task, to approach four different trees during a human-robot field survey. 
The robot was capable of localizing, orientating towards, and reaching all commanded locations in an average of less than $34~s$ for approximately up to $12~m$ away-placed targets. 
Each target was reached with Root Mean Squared Error (RMSE) in the x-axis at $0.41~m$ and y-axis at $0.67~m$ compared with the created AR target goals (namely, the tree inspection locations), staying in a safe position away from the tree, while also reaching the closest requested point.

\section{Conclusion}
\label{sec:conclusion}
This work demonstrates the potential of integrating AR with mobile robotics to support precision agriculture. 
By leveraging open-source computational tools and off-the-shelf hardware, the developed framework provides real-time data input and control output through a virtual environment enabled by an AR headset interface. 
The system's capabilities are showcased in a case study application, termed ``Holoagro,'' where growers or technicians can interact with a legged robot via an AR headset and a custom UI. 
This interaction supports multiple modes of operation. 
First, it affords teleoperation for scene understanding (e.g., looking for irrigation line leaks or other field-related inspection tasks).
It also allows a user to request on-demand and real-time status updates of specific areas of the field (e.g., growth and vigor of trees), and call a robot ``assistant" to make those measurements and update the underlying database, autonomously. 
The latter functionality is enabled by a developed local navigation scheme which enables navigation in fused virtual and physical environments. 

Future work aims to include multimodal sensing modalities adapted to different robotic platforms. 
We aim to expand the framework to encompass various robotic platforms, such as aerial and wheeled robots, enabling more comprehensive human-robot collaboration applications in precision agriculture. 
Alternative navigation and localization methods for multi-robot systems will also be explored to further improve the system’s robustness and versatility. 
Additionally, the system’s capabilities can be extended to other scenarios, such as fire management or search and rescue. 
Overall, this work represents an exciting step toward integrating AR and robotics in agriculture, providing practical and readily implementable solutions for real-time data management, and teleoperated and autonomous navigation of field robots, achieved through the use of custom UIs and open-source augmented reality tools.

\bibliographystyle{ieeetr}
\bibliography{root}

\end{document}